\documentclass[preprint,12pt,number]{elsarticle}

\usepackage{ifpdf}
\usepackage{import}
\usepackage{amsmath,amssymb,amsfonts}
\usepackage{graphicx}
\usepackage{makecell,multirow,booktabs,float}
\usepackage[T1]{fontenc}
\usepackage{makecell,multirow,float}
\usepackage[hyphens]{url}     
\usepackage[hidelinks]{hyperref} 
\usepackage[ruled,vlined]{algorithm2e}
\usepackage{tabularx}

\begin{document}

\begin{frontmatter}

\title{Distributed Hierarchical Temporal Memory with Shared Associative Memory for Cross-Entity Preemptive Warning}

\author[usf]{Pavia Bera\corref{cor1}}
\ead{paviabera@usf.edu}
\author[usf]{Jennifer Adorno}
\author[usf]{Sanjukta Bhanja}
\address[usf]{University of South Florida, Tampa, FL, USA}
\cortext[cor1]{Corresponding author.}

\begin{abstract}
Anomaly detection in multivariate time series remains a critical challenge in large-scale distributed systems, where related entities may exhibit transferable precursor behavior prior to anomaly onset. Existing methods typically operate independently on each data stream and therefore remain fundamentally reactive. To address this limitation, we introduce Distributed Hierarchical Temporal Memory (D-HTM), a neuromorphic framework that enables cross-entity preemptive warning through a Shared Associative Memory (SAM).

D-HTM combines a Spatial Pooler (SP) that projects observations into a common Sparse Distributed Representation (SDR) space, Temporal Memory (TM) modules that learn entity-specific dynamics online, and a Shared Associative Memory that stores recurring pre-anomaly signatures. By reusing precursor knowledge across related entities, D-HTM can issue warnings prior to local anomaly onset while preserving HTM's online learning capabilities.

We evaluate D-HTM on the Server Machine Dataset (SMD), the Soil Moisture Active Passive (SMAP) dataset, the Mars Science Laboratory (MSL) dataset, and a synthetic cascade benchmark designed to isolate precursor transfer. Experimental results demonstrate effective cross-entity warning propagation while maintaining competitive reactive anomaly detection performance. Across the real-world datasets, D-HTM provides an average warning lead time of 8.1 samples prior to anomaly onset.

These findings demonstrate that transferable precursor structure can emerge within a shared SDR space and be reused for preemptive warning generation, extending HTM beyond isolated reactive detection toward distributed predictive reasoning.
\end{abstract}


\end{frontmatter}

\section{Introduction}
\label{sec:intro}

The explosive growth in data generation from internet-connected devices
has placed unprecedented demands on intelligent systems to process
information efficiently and in real time. While traditional neural
networks and other artificial intelligence (AI) algorithms have achieved
remarkable success in domains such as image recognition and natural
language processing, they face significant limitations when adapting to
continuous, dynamic streams of unlabeled data~\cite{laird2017standard}.
These systems typically rely on batch learning, requiring extensive
training data, computational resources, and retraining when the input
distribution changes. Moreover, their architectures are not inherently
designed for online inference and adaptation, which are critical
requirements in applications such as anomaly detection, robotics, and
streaming data analysis.

In contrast, Hierarchical Temporal Memory (HTM) is a biologically
inspired framework modeled on the structure and function of the human
neocortex. HTM learns continuously from streaming data by forming sparse
distributed representations and temporal associations without requiring
offline retraining~\cite{hawkins2016neurons,ahmad2016neurons}. Its
online learning capability has enabled applications in anomaly detection,
robotics, forecasting, medical imaging, wafer inspection, biometric
recognition, and environmental
monitoring~\cite{lavin2015evaluating,zhou2018hierarchical,adam2018wafer,james2017htm,neubert2018sequence,micheletto2018using,osegi2018using,zyarah2020end}.

\begin{figure}[t]
\centering
\includegraphics[width=1\columnwidth]{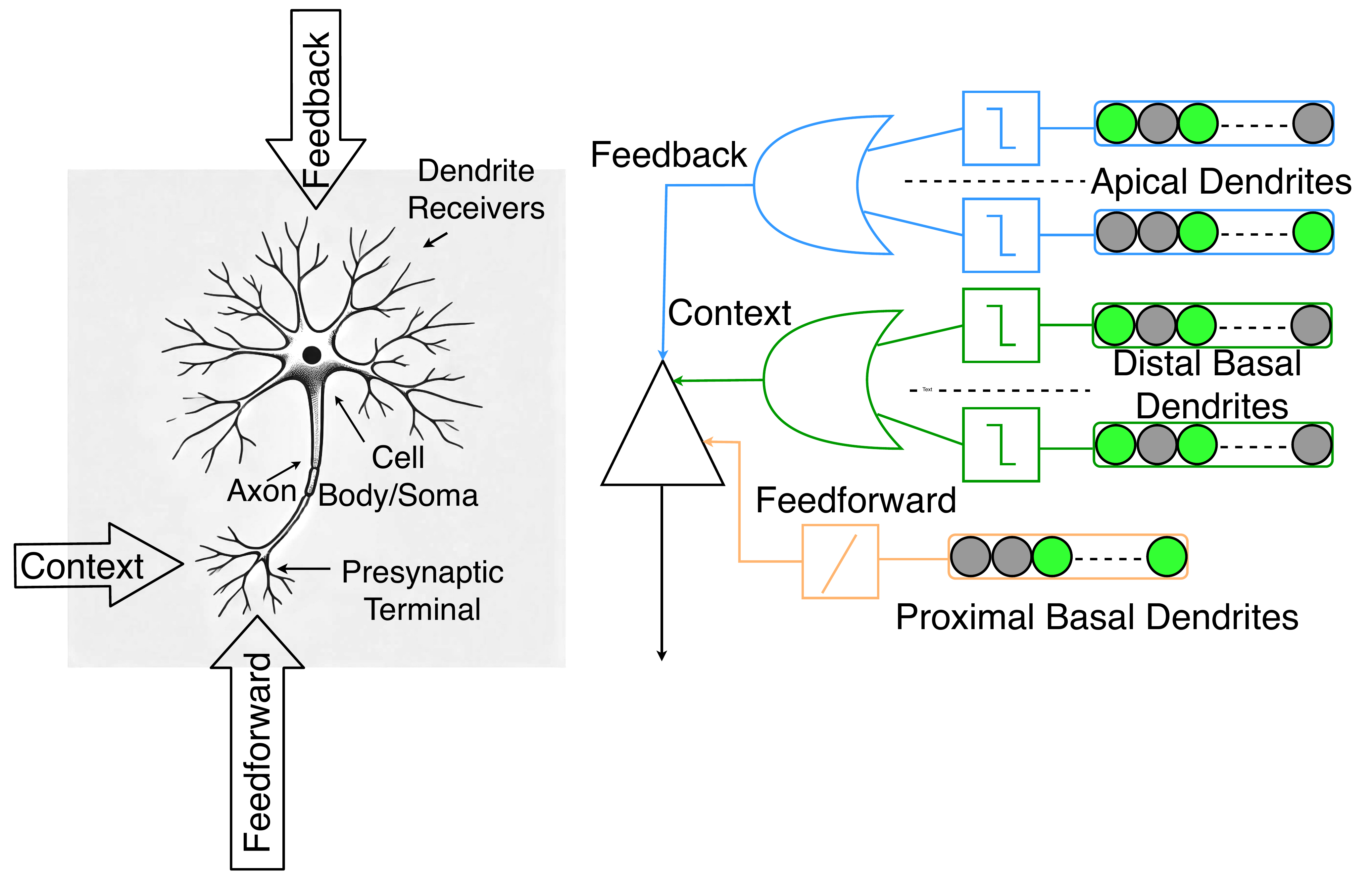}
\vspace{-4ex}
\caption{Comparison of biological and HTM neurons, highlighting feedback,
feedforward, and contextual connections in both systems.}
\vspace{-3ex}
\label{fig:Biological_HTM}
\end{figure}

At the core of HTM is the Temporal Memory (TM), which models key
processing mechanisms of pyramidal neurons within the
neocortex~\cite{zyarah2020end,bautista2020matlabhtm}, as illustrated in
Fig.~\ref{fig:Biological_HTM}. Through interactions among proximal,
distal, and apical dendrites, TM learns predictive associations between
successive sparse activation patterns, enabling continuous temporal
learning and anomaly detection without offline
retraining~\cite{hawkins2016neurons,bautista2020matlabhtm,harris2015neocortical}.

Despite these strengths, HTM deployments in large-scale distributed
environments remain fundamentally isolated. Each HTM agent learns only
from its own local observations, even though anomalies in real-world
systems frequently propagate across related
entities~\cite{audibert2020usad}. Consequently, existing anomaly
detection methods, including state-of-the-art approaches such as
OmniAnomaly~\cite{su2019omnianomaly}, process each data stream
independently and cannot exploit knowledge acquired elsewhere in the
system. Anomaly detection therefore remains largely reactive, with each
agent responding only after anomalous behavior becomes observable within
its own data stream.

To address this limitation, we propose Distributed Hierarchical Temporal
Memory (D-HTM), a framework that augments distributed HTM agents with a
Shared Associative Memory (SAM). SAM stores recurring precursor
activation patterns observed before anomaly onset and enables this
knowledge to be shared across agents. When a local temporal pattern
resembles a previously stored precursor, the retrieved memory contributes
evidence toward a pre-emptive warning before local anomaly scores reach
detection thresholds, allowing experience acquired by one entity to
improve situational awareness across the network.

Unlike centralized monitoring architectures, D-HTM preserves HTM's
online learning capability by allowing each agent to learn independently
while asynchronously querying and updating the shared associative memory.
No global synchronization or model retraining is required.

The main contributions of this paper are as follows:

\begin{itemize}

\item We introduce a Shared Associative Memory (SAM) that stores and
retrieves recurring pre-anomaly sparse activation patterns, enabling
transferable precursor knowledge across distributed HTM agents.

\item We propose an inference-time cross-entity warning mechanism in
which retrieved precursor memories generate pre-emptive warnings before
local anomaly scores reach detection thresholds.

\item We evaluate the proposed framework on the SMD, SMAP, and MSL
benchmarks, demonstrating competitive reactive anomaly detection while
showing that shared associative memory enables effective cross-entity
warning propagation.

\end{itemize}

The remainder of this paper is organized as follows.
Section~\ref{sec:rel} reviews related work on HTM, distributed anomaly
detection, and neuromorphic computing.
Section~\ref{sec:background} introduces the HTM components relevant to
this work.
Section~\ref{sec:framework} presents the proposed Distributed HTM
framework and the Shared Associative Memory.
Section~\ref{sec:experiments} reports the experimental evaluation, and
Section~\ref{sec:conclusion} concludes the paper.
 
\section{Related Work}
\label{sec:rel}

Anomaly detection in distributed multivariate systems has been studied
from several perspectives, including distributed learning, biologically
inspired computing, and proactive anomaly detection. This section reviews
the most relevant work and highlights the gap addressed by the proposed
Shared Associative Memory (SAM) framework.

\textbf{Distributed and cooperative anomaly detection:}
Multi-agent learning has been widely investigated as a means of improving
scalability and robustness in complex monitoring
environments~\cite{foerster_ma_rl,rusu_progressive}. Classical
distributed systems rely on centralized controllers or shared replay
buffers, while more recent biologically inspired approaches emphasize
decentralized coordination in which agents learn locally and periodically
exchange information~\cite{vashist_dmem}. In temporal sequence modeling,
distributed learning has primarily been built upon recurrent neural
networks and attention-based
architectures~\cite{transformer_distributed,rnn_multiagent}, which
require offline gradient-based optimization and are not inherently
designed for continual online adaptation.

Federated learning (FL) has emerged as the dominant paradigm for
distributed multivariate anomaly detection, allowing local models to be
trained independently while exchanging only model
updates~\cite{mcmahan2017communication}. Recent work has extended this
idea through multi-task federated
learning~\cite{hao2024sadmc}, hypernetwork-based
aggregation~\cite{hao2025ufedhy}, IIoT intrusion
detection~\cite{zheng2025maslstm}, and LLM-based multi-agent
coordination~\cite{yang2025adagent,lemad2025}. Complementing these,
Audibert \emph{et al.}~\cite{audibert2020usad} demonstrated that
anomalies in complex distributed systems rarely occur in isolation, with
failure signatures frequently propagating across related machines.
Similarly, attempts to scale HTM to distributed environments have relied
primarily on centralized coordination or static
partitioning~\cite{numenta_whitepaper}, without providing a mechanism for
sharing learned temporal knowledge across agents.

Despite their differences, existing distributed approaches communicate
primarily during \emph{training} by exchanging gradients, model
parameters, or task-level information. During online inference, each
agent operates independently, with no mechanism to propagate learned
precursor patterns to neighboring agents before local anomaly onset. This
distinction fundamentally separates the proposed inference-time warning
mechanism from existing distributed learning approaches.

\textbf{Biologically inspired and neuromorphic approaches:}
Hierarchical Temporal Memory (HTM) is a biologically inspired framework
that models neocortical computation using Sparse Distributed
Representations (SDRs) and online temporal
learning~\cite{numenta_whitepaper,ahmad2016why}. Its ability to learn
continuously without retraining has made HTM well suited for streaming
analytics and anomaly detection~\cite{bera2025hahtm}. Complementary
neuromorphic approaches, including spiking neural networks and
specialized hardware such as
Loihi~\cite{shen2025cgsnn,gatti2025cxray,davies_loihi}, likewise exploit
sparse asynchronous computation for continual temporal inference.
However, these biologically inspired systems remain predominantly
single-agent and do not support sharing learned anomaly patterns across
independently operating agents in real time.

\textbf{Proactive and pre-emptive anomaly detection:}
A complementary line of research focuses on detecting anomalies before
they fully manifest. Early work by Thottan and
Ji~\cite{thottan1998proactive} introduced distributed proactive
monitoring using intelligent agents, while RePAD~\cite{lee2020repad}
extended this concept through online LSTM-based prediction for streaming
time series. More recently, forecasting-based proactive
methods~\cite{jeon2025proactive}, factor-graph reasoning for intrusion
detection~\cite{cao2015preemptive}, and early-detection
benchmarks~\cite{ltg2025early} have demonstrated that meaningful
precursor information can often be identified before anomaly onset.

Nevertheless, these approaches remain fundamentally single-stream. Early
warnings are generated solely from observations within the local data
stream, and no mechanism exists for transferring learned precursor
knowledge across related entities. This limitation also applies to
federated methods, where communication is restricted to model updates and
does not occur during online inference.

\textbf{Benchmarks and evaluation:}
The Server Machine Dataset (SMD)~\cite{su2019omnianomaly}, together with
the SMAP and MSL telemetry benchmarks, has become the standard evaluation
suite for multivariate time-series anomaly detection. OmniAnomaly,
introduced alongside these datasets, remains one of the most widely used
reactive baselines for anomaly
detection~\cite{su2019omnianomaly}. Subsequent methods, including
InterFusion~\cite{li2021interfusion} and
StackVAE~\cite{stackvae2022}, have further improved reactive detection
performance under the standard evaluation protocols. These methods,
however, are designed for single-entity anomaly detection and therefore
provide the reactive baselines against which the proposed cross-entity
warning framework is evaluated.

Table~\ref{tab:related} summarizes the key architectural differences
between representative anomaly detection approaches and the proposed SAM
framework. The comparison emphasizes the properties most relevant to this
work, including online learning capability, per-machine deployment,
cross-agent communication, and support for pre-emptive warning
generation.

\begin{table*}[!ht]
\centering
\caption{Comparison of distributed and cooperative anomaly detection methods.
\textbf{Bold} highlights the differentiating properties of the proposed
SAM framework. $\checkmark$~=~yes; $\times$~=~no.}
\label{tab:related}
\renewcommand{\arraystretch}{1.3}
\resizebox{\textwidth}{!}{%
\begin{tabular}{llcclc}
\toprule
\textbf{Method} &
\textbf{Paradigm} &
\textbf{\shortstack[c]{Per-\\mach.}} &
\textbf{\shortstack[c]{On-\\line}} &
\textbf{Cross-agent communication} &
\textbf{\shortstack[c]{Pre-\\empt.}} \\
\midrule
OmniAnomaly~\cite{su2019omnianomaly}  & Stochastic RNN (VAE+NF)       & $\times$     & $\times$     & None                              & $\times$     \\
InterFusion~\cite{li2021interfusion}  & Hierarchical VAE              & $\times$     & $\times$     & None                              & $\times$     \\
StackVAE~\cite{stackvae2022}          & Stacked VAE + GNN             & $\times$     & $\times$     & None                              & $\times$     \\
USAD~\cite{audibert2020usad}          & Dual-AE adversarial           & $\times$     & $\times$     & None                              & $\times$     \\
RePAD~\cite{lee2020repad}             & LSTM online prediction        & $\times$     & $\checkmark$ & None                              & $\checkmark$ \\
Jeon et al.~\cite{jeon2025proactive}  & Forecasting proactive         & $\times$     & $\times$     & None                              & $\checkmark$ \\
uFedHy~\cite{hao2025ufedhy}           & Federated hypernetwork        & $\checkmark$ & $\times$     & Gradient aggregation (train-time) & $\times$     \\
SADMC~\cite{hao2024sadmc}             & Multi-task federated FL       & $\checkmark$ & $\times$     & Gradient aggregation (train-time) & $\times$     \\
MAS-LSTM~\cite{zheng2025maslstm}      & Decentralized LSTM + GSP      & $\checkmark$ & $\times$     & None (parallel agents)            & $\times$     \\
AD-AGENT~\cite{yang2025adagent}       & LLM pipeline orchestration    & $\times$     & $\times$     & Workflow memory (train-time)      & $\times$     \\
LEMAD~\cite{lemad2025}                & Hierarchical LLM-MAS          & $\checkmark$ & $\times$     & Task delegation (train-time)      & $\times$     \\
\midrule
\textbf{SAM (ours)} &
\textbf{HTM + Shared Associative Memory} &
$\checkmark$ &
$\checkmark$ &
\textbf{Associative precursor memories (inference-time)} &
$\checkmark$ \\
\bottomrule
\end{tabular}}
\end{table*}

As summarized in Table~\ref{tab:related}, existing distributed anomaly
detection methods exchange gradients, model parameters, or task-level
information during training, while proactive methods generate warnings
only from the local data stream. In contrast, the proposed SAM framework
introduces inference-time communication through retrieval of previously
learned precursor memories. Rather than sharing model parameters, agents
share transferable temporal knowledge encoded as sparse precursor
patterns, enabling early warnings to propagate across related entities
before local anomaly onset.

```latex id="a7mw0s"
\section{Background}
\label{sec:background}

\begin{figure}[t]
\centering
\includegraphics[width=0.9\linewidth]{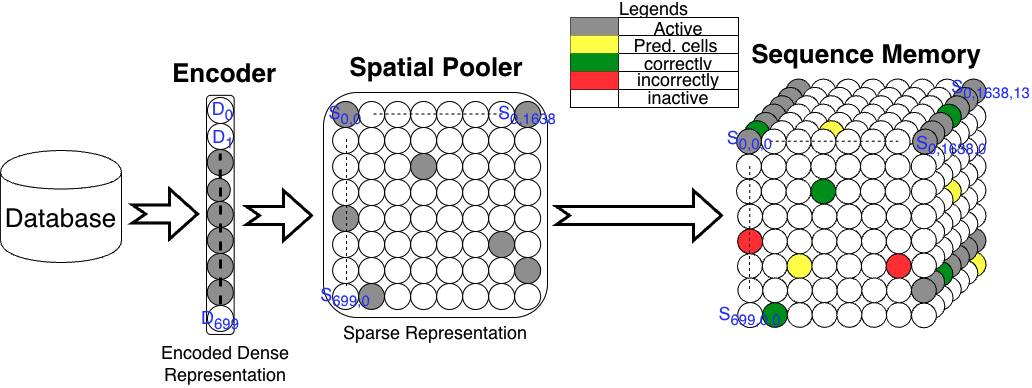}
\caption{Overview of the Hierarchical Temporal Memory (HTM) architecture,
including the encoder, Spatial Pooler (SP), and Temporal Memory (TM).}
\label{fig:htm_pipeline}
\end{figure}

Hierarchical Temporal Memory (HTM) is a biologically inspired machine
learning framework modeled after the mammalian neocortex. Unlike
conventional deep learning methods that require offline training and
periodic retraining, HTM continuously learns from streaming data while
simultaneously performing inference. As illustrated in
Fig.~\ref{fig:htm_pipeline}, the HTM pipeline consists of three
components: an encoder, a Spatial Pooler (SP), and a Temporal Memory
(TM), which together transform raw observations into sparse
representations and learn their temporal relationships.

The encoder converts raw numerical, categorical, or spatial inputs into
Sparse Distributed Representations (SDRs), high-dimensional binary
vectors containing only a small fraction of active bits. SDRs preserve
semantic similarity through overlapping active bits while remaining
robust to noise and small perturbations. More importantly for this work,
their sparse overlap statistics make random matches exceedingly unlikely,
providing a reliable basis for overlap-based similarity matching across
entities.

The Spatial Pooler maps encoder-generated SDRs into a fixed population of
sparse mini-columns using k-Winners-Take-All inhibition and local
Hebbian-like learning. Similar inputs therefore activate overlapping
column subsets, producing a stable sparse representation that preserves
input similarity while reducing sensitivity to noise. In the proposed
framework, this property enables different entities to share a common
representation space suitable for cross-entity comparison.

Temporal Memory extends HTM into the temporal domain by learning
transitions between successive Spatial Pooler activations. By forming
predictive associations between sparse activation patterns, TM performs
continuous sequence learning and produces anomaly scores from prediction
errors without requiring offline retraining.

The proposed framework relies on three key properties of HTM: (i) SDRs
provide robust sparse representations with reliable overlap statistics,
(ii) the Spatial Pooler maps similar inputs to comparable sparse
activations, and (iii) Temporal Memory continuously learns temporal
structure while detecting anomaly onset through prediction errors.
Together, these properties enable the storage, retrieval, and transfer
of recurring precursor patterns across distributed entities, forming the
foundation of the Shared Associative Memory (SAM) framework introduced in
the following section.

\section{Distributed HTM Framework (D-HTM)}
\label{sec:framework}

\subsection{Cross-Entity Representation Alignment}
\label{subsec:alignment}

A fundamental challenge in cross-entity anomaly warning is that similar
physical events may appear substantially different across monitored
entities. For example, a sudden increase in CPU utilization may represent
a severe deviation on a machine that normally operates at 20\% load,
while the same increase may be routine for another machine operating at
70\% load. Direct comparison of raw measurements therefore suffers from
operating-point bias and can obscure shared precursor patterns.

To reduce this bias, each feature is independently normalized using
z-score normalization. For entity $i$ and feature $j$, the normalized
value is

\begin{equation}
z_{i,j}(t)=\frac{x_{i,j}(t)-\mu_{i,j}}{\sigma_{i,j}},
\end{equation}

where $\mu_{i,j}$ and $\sigma_{i,j}$ are computed from the training
portion of the corresponding entity's data and remain fixed throughout
evaluation. The resulting normalized features describe deviations
relative to each entity's normal operating regime rather than absolute
measurement values, making similar behaviors more directly comparable.

Normalization alone, however, is insufficient. If each entity trains an
independent Spatial Pooler (SP), similar normalized inputs may still
activate different SDR columns because the learned column assignments
evolve independently. Consequently, overlap between SDRs from different
entities would no longer reflect semantic similarity.

To overcome this limitation, all entities share a single Spatial Pooler.
The shared SP is trained once using representative historical data and
then frozen, ensuring that comparable input patterns generate
overlapping sparse representations regardless of their source entity.
Together, z-score normalization and the shared SP establish a common SDR
space that enables reliable overlap-based comparison and forms the
foundation of the Shared Associative Memory.

\subsection{System Architecture}
\label{subsec:architecture}

\begin{figure}[t]
\centering
\includegraphics[width=0.9\linewidth]{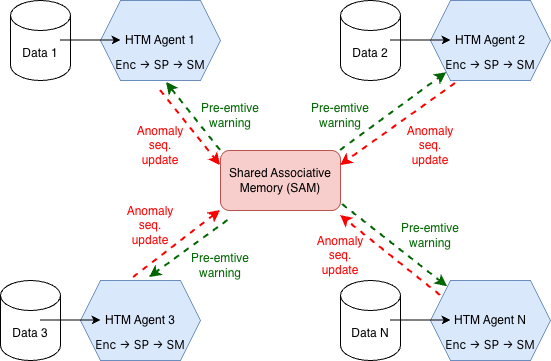}
\caption{Overview of the Distributed HTM (D-HTM) framework. Each entity
processes a local data stream through z-score normalization, an encoder,
a shared Spatial Pooler (SP), and a local Temporal Memory (TM).
Precursor memories extracted before anomaly onset are stored in the
Shared Associative Memory (SAM), enabling cross-entity warning
propagation through memory retrieval.}
\label{fig:ma_architecture}
\end{figure}

Having established a common representation space, we now describe the
overall Distributed HTM (D-HTM) framework. D-HTM consists of $N$
distributed HTM agents that share a Spatial Pooler and a Shared
Associative Memory (SAM), while maintaining independent Temporal Memory
(TM) modules for local sequence learning. An overview of the architecture
is shown in Fig.~\ref{fig:ma_architecture}.

For each entity $i$, an incoming multivariate data stream $X_i(t)$ is
normalized, encoded into Sparse Distributed Representations (SDRs), and
processed by the shared Spatial Pooler. The resulting sparse
representation is supplied to the local Temporal Memory, which learns
entity-specific temporal dynamics and computes anomaly scores from
prediction errors.

When a local anomaly onset is detected, the preceding $L$ Spatial
Pooler activations are extracted and stored in SAM as a precursor
memory,

\begin{equation}
M_j=
\left\{
s_{t-L},
s_{t-L+1},
\dots,
s_{t-1}
\right\},
\end{equation}

where each $s_t$ denotes the active Spatial Pooler columns at time $t$.
The lookback horizon $L$ determines the amount of temporal context stored
before anomaly onset and is treated as a dataset-dependent parameter.
Its influence on warning performance is evaluated in
Section~\ref{sec:robustness}.

Overall, D-HTM combines localized temporal learning with a shared
repository of precursor memories. Each agent continues to learn
independently from its own data stream while contributing to and querying
SAM during online operation. The memory formation, retrieval, and update
procedures are described in Section~\ref{subsec:SAM}.

\subsection{Shared Spatial Pooler}
\label{subsec:Shared SP}

To enable cross-entity comparison, D-HTM replaces the conventional
per-entity Spatial Pooler (SP) with a single Shared Spatial Pooler.

The Shared SP is trained once using the combined normal operating data
from all monitored entities during an offline initialization stage.
After convergence, its parameters are frozen and reused by every HTM
agent throughout deployment, while each agent maintains an independent
Temporal Memory for online sequence learning.

Because all agents share the same Spatial Pooler, similar normalized
inputs are projected into comparable Sparse Distributed Representations
(SDRs) regardless of their source entity. Consequently, precursor
patterns generated by different entities can be compared directly using
sparse overlap without requiring alignment of raw feature values or
entity-specific SP representations.

Only the feature representation is shared. Temporal learning remains
localized within each Temporal Memory, allowing every entity to adapt to
its own sequential dynamics while benefiting from a common sparse
representation for cross-entity precursor retrieval.

\subsection{Shared Associative Memory and Warning Propagation}
\label{subsec:SAM}

\begin{figure*}[!t]
\centering
\includegraphics[width=\textwidth]{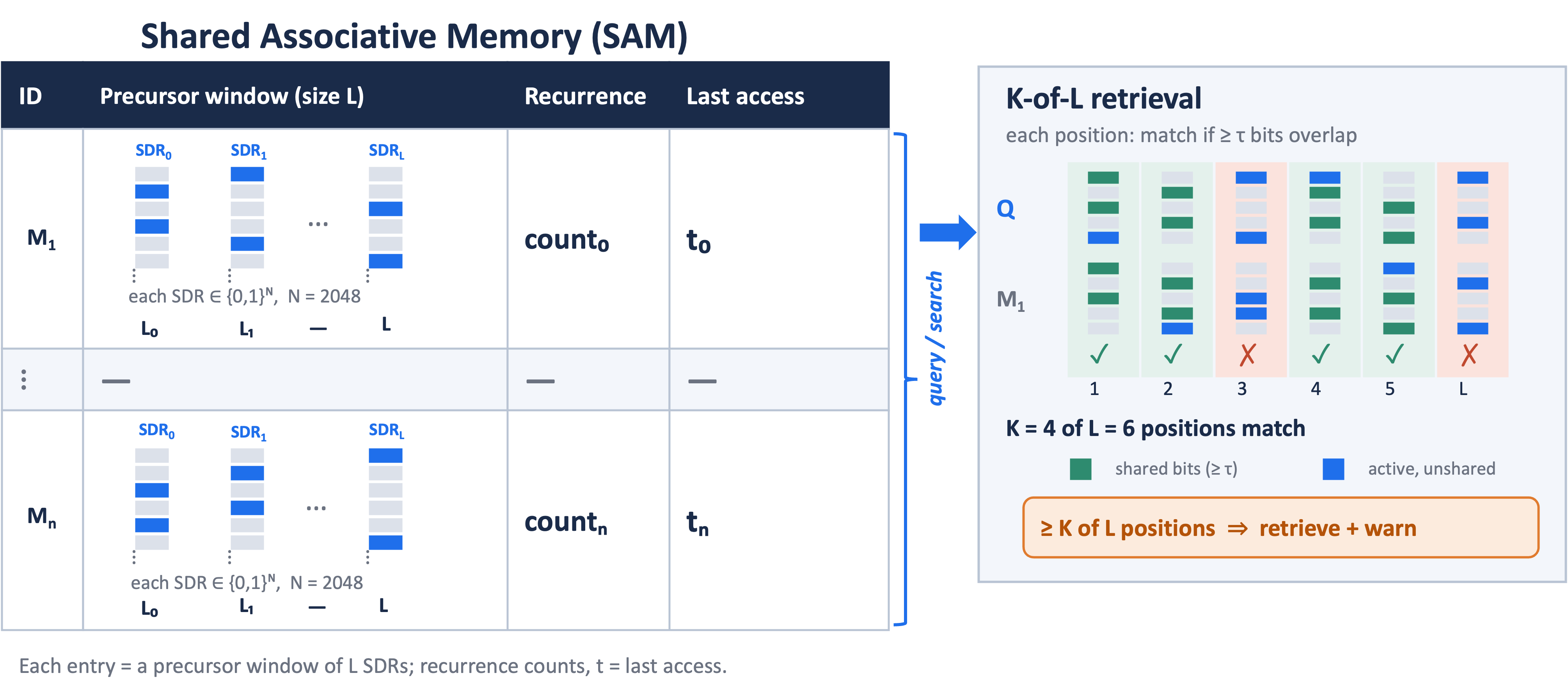}
\caption{Structure of the Shared Associative Memory (SAM) and its retrieval process. Each memory entry stores a precursor window of $L$ Spatial Pooler SDRs together with its recurrence count ($n$) and last-access time ($t$). During inference, the current query window is compared against stored memories using K-of-$L$ sparse overlap matching. A warning is issued when at least $K$ of the $L$ corresponding SDRs satisfy the overlap threshold $\tau$.}
\label{fig:sam}
\end{figure*}

Once a common representation has been established by the Shared Spatial
Pooler, recurring precursor patterns can be stored in a shared memory and
reused across entities. This functionality is provided by the Shared Associative Memory (SAM),
illustrated in Fig.~\ref{fig:sam}, which stores activation patterns that
have previously preceded anomaly onset.

Each memory entry consists of a precursor activation window together with
metadata describing its usage history,

\begin{equation}
M_j=(W_j,n_j,t_j),
\end{equation}

where
$W_j=\{w_j^0,w_j^1,\ldots,w_j^{L-1}\}$
denotes the stored precursor window,
$n_j$ records the recurrence count, and
$t_j$ stores the most recent access time.
The recurrence count serves as a confidence measure, reflecting how
frequently the same precursor has been observed across all monitored
entities.

Whenever a local anomaly onset is detected, the preceding $L$ Spatial
Pooler activation patterns are extracted and compared with existing
entries in SAM. If a sufficiently similar memory already exists, its
recurrence count is incremented; otherwise, a new memory entry is
created. This reinforcement mechanism prevents duplicate memories while
naturally increasing the confidence of frequently recurring precursor
patterns.

During inference, each entity continuously constructs a query window from
its most recent Spatial Pooler activations,

\begin{equation}
Q_t=\{q_{t-L},q_{t-L+1},\ldots,q_{t-1}\}.
\end{equation}

For every stored memory, corresponding SDRs are compared position by
position. A positional match is declared whenever

\begin{equation}
|q^p\cap w_j^p|\ge\tau,
\label{eq:overlap}
\end{equation}

where $q^p$ denotes the query SDR, $w_j^p$ denotes the stored SDR at the
same position, and $\tau$ specifies the minimum overlap required for two
SDRs to be considered similar. In this work, $\tau$ remains fixed across
all experiments.

A precursor memory is retrieved when at least $K$ of the $L$ positions
satisfy Eq.~(\ref{eq:overlap}),

\begin{equation}
\mathrm{warning}(t)=
\max_j
\left(
\sum_{p=0}^{L-1}
\mathbf{1}
\left[
|q^p\cap w_j^p|\ge\tau
\right]
\right)
\ge K,
\label{eq:kofl}
\end{equation}

where $K$ controls the minimum number of matching positions required to
issue a warning. Larger values of $K$ demand stronger temporal agreement,
while smaller values permit more tolerant retrieval.

Retrieved memories contribute evidence toward a pre-emptive warning
before the local anomaly detector signals an anomaly. The recurrence
count reflects the historical reliability of each precursor, while the
last-access timestamp supports memory management. When the memory reaches
its capacity, entries with low recurrence counts and long periods of
inactivity are preferentially removed, allowing frequently reused
precursor memories to remain while obsolete patterns are gradually
forgotten.

Algorithm~\ref{alg:SAM} summarizes the complete online operation of SAM.
At every timestep, memory retrieval is performed before any new memory is
inserted, ensuring that warnings are generated exclusively from precursor
patterns learned during previous anomaly events.

\begin{algorithm}[!t]
\caption{Shared Associative Memory (SAM) Retrieval and Update}
\label{alg:SAM}
\KwIn{Current SDR $s(t)$, lookback horizon $L$, overlap threshold $\tau$, match threshold $K$, shared memory $\mathcal{M}$}
\KwOut{Pre-emptive warning}

Form query window
$Q_t=\{s(t-L),\ldots,s(t-1)\}$\;

\ForEach{memory $M_j=(W_j,n_j,t_j)\in\mathcal{M}$}{
    Compute the number of matching positions between $Q_t$ and $W_j$\;

    \If{at least $K$ positions satisfy $|q^p\cap w_j^p|\ge\tau$}{
        Issue warning\;
        Update last-access timestamp $t_j\leftarrow t$\;
    }
}

\If{local HTM detects anomaly onset}{
    Extract precursor window
    $W=\{s(t-L),\ldots,s(t-1)\}$\;

    \If{$W$ matches an existing memory}{
        Increment recurrence count
        $n_j\leftarrow n_j+1$\;
        Update last-access timestamp
        $t_j\leftarrow t$\;
    }
    \Else{
        Insert new memory
        $(W,1,t)$ into $\mathcal{M}$\;
    }
}

\If{memory capacity exceeded}{
    Remove memories with the lowest recurrence count and oldest
    last-access time\;
}
\end{algorithm}

Because $L$ is measured in samples rather than physical time, the
effective precursor horizon depends on the sampling interval and failure
development characteristics of each dataset. Consequently, the optimal
value of $L$ is dataset dependent and is investigated in
Section~\ref{sec:robustness}.

\section{Experiments and Discussion}
\label{sec:experiments}

\subsection{Datasets and Preprocessing}
\label{sec:datasets}

We evaluate the proposed framework on three established multivariate
anomaly detection benchmarks—SMD~\cite{su2019omnianomaly},
SMAP~\cite{telemanom}, and MSL~\cite{telemanom}—together with a
synthetic cascade dataset designed to evaluate controlled cross-entity
warning propagation. SMD contains monitoring traces from multiple
production servers, while SMAP and MSL comprise NASA spacecraft
telemetry collected from multiple related channels. The synthetic
dataset provides a controlled environment in which precursor trajectories
are intentionally shared across entities, enabling evaluation of the
upper bound of transferable warning performance. Dataset statistics are
summarized in Table~\ref{tab:datasets}.

\begin{table}[h]
\centering
\setlength{\tabcolsep}{5pt}
\caption{Dataset summary. Steps are reported per entity; anomaly statistics
are computed over the test split only.}
\label{tab:datasets}
\begin{tabular}{lcccccc}
\toprule
Dataset &
Entities &
Features &
\shortstack{Train\\Steps} &
\shortstack{Test\\Steps} &
\shortstack{Anomaly\\Events} \\
\midrule
SMD~\cite{su2019omnianomaly}
& 28 & 38 & $\sim$28,000 & $\sim$28,000 & 327 \\
SMAP~\cite{telemanom}
& 54$^{a}$ & 25 & 312--2,881 & 4,453--8,640 & 68 \\
MSL~\cite{telemanom}
& 27 & 55 & 439--4,308 & 1,096--6,100 & 36 \\
Synthetic (ours)
& 6 & 15 & 5,000 & 5,000 & 24 \\
\bottomrule
\end{tabular}
\end{table}

Preprocessing is performed independently for each entity to prevent
information leakage. All feature statistics are computed exclusively
from the training split and remain fixed throughout evaluation. Each
feature is z-score normalized using its training-set mean and standard
deviation. Normalization is applied per machine for SMD, per telemetry
channel for SMAP and MSL, and per node for the synthetic dataset.
Features with near-zero variance ($\sigma<10^{-8}$) are mapped to zero,
and normalized values are clipped to $[-10,10]$ prior to encoding.

For SMD, we follow the standard train/test protocol of
OmniAnomaly~\cite{su2019omnianomaly}. The first half of each machine
trace is used for training and the second half for testing. For SMAP and MSL, we use the official Telemanom~\cite{telemanom} train/test splits. Three SMD machines (machine-2-4, machine-3-1, and machine-3-3) are
excluded because their normalized pre-anomaly profiles become
numerically degenerate, producing unstable similarity estimates.

\subsection{Implementation Details}
\label{sec:implementation}

Unless otherwise stated, identical encoder, Spatial Pooler (SP), and
Temporal Memory (TM) configurations are used across all real-world
datasets to preserve consistent Sparse Distributed Representation (SDR)
statistics during cross-entity retrieval. The complete implementation
parameters are summarized in Table~\ref{tab:implementation}.

Each scalar feature is encoded using a 512-bit
\texttt{ScalarEncoder} with 41 active bits. Per-feature encodings are
concatenated before Spatial Pooling, yielding an expected random SDR
overlap of approximately 1.64 bits, which serves as the effective noise
floor for overlap-based retrieval.

The shared SP contains 1,024 mini-columns with a target sparsity of
4\% ($\approx41$ active columns) and uses global inhibition. It is
trained offline for one epoch using shuffled training data and then
frozen during inference, preserving consistent SDR semantics across all
entities. Each entity maintains an independent TM that continues online
learning throughout inference. Prediction errors are converted into
anomaly scores and smoothed using a rolling mean prior to thresholding.

Four TM hyperparameters are optimized independently for each dataset
using Optuna~\cite{akiba2019optuna} with 40 optimization trials,
maximizing the mean point-adjust F1 on the validation set. The search
includes \texttt{boostStrength},
\texttt{cellsPerColumn},
\texttt{activationThreshold}, and
\texttt{predictedSegmentDecrement}. Hyperparameter optimization is
performed on the USF CIRCE HPC cluster.

For all datasets, the final 20\% of the training split is reserved for
validation. Hyperparameter optimization and lookback-horizon selection
are performed exclusively on this validation subset. After parameter
selection, the model is retrained using the complete training data and
evaluated once on the held-out test set. Test labels are never used
during parameter tuning or lookback selection.

Memory insertion follows a strictly online protocol. At each timestep,
SAM first queries the shared memory using only information available up
to the current time. If the local HTM detector subsequently identifies
an anomaly onset, the preceding $L$ Spatial Pooler activations are
stored as a precursor memory. Ground-truth anomaly labels are used
exclusively for evaluation and are never available during memory
formation or retrieval, ensuring a leakage-free online evaluation.

We compare D-HTM against three baselines.
OmniAnomaly~\cite{su2019omnianomaly} serves as the primary deep-learning
baseline following the standard SMD/SMAP/MSL evaluation protocol. A
single-entity HTM baseline isolates the contribution of the shared SP by
running HTM independently for each entity without cross-entity
communication. Finally, an encoder-only D-HTM ablation replaces SP
outputs with raw \texttt{ScalarEncoder} bit arrays to evaluate the
importance of learned sparse representations for transferable precursor
matching.

Performance is evaluated using Precision, Recall, and F1 under the
point-adjust protocol~\cite{su2019omnianomaly}. For cross-entity warning
evaluation, we additionally report Warning Precision, Event Recall,
Mean Lead Time, and warning confidence.

The benchmark datasets contain relatively few anomaly events, resulting
in far fewer precursor memories than the configured SAM capacity.
Consequently, memory eviction is rarely triggered during the reported
experiments. The recurrence- and recency-based replacement policy is
primarily intended for long-running deployments where memory growth
eventually exceeds the available capacity.

\begin{table}[!t]
\footnotesize
\setlength{\tabcolsep}{3pt}
\caption{Implementation parameters used across experiments.}
\label{tab:implementation}
\begin{tabular}{p{0.48\columnwidth}p{0.42\columnwidth}}
\hline
Parameter & Value \\
\hline
Encoder size & 512 bits \\
Encoder active bits & 41 \\
Encoder input range & [-10,10] (z-score) \\
Expected random SDR overlap & 1.64 bits \\
\hline
SP columns & 1,024 \\
SP sparsity & 4\% \\
SP inhibition & Global \\
SP training & 1 offline epoch \\
SP inference & Frozen \\
\texttt{potentialPct} & 0.85 \\
\texttt{synPermInactiveDec} & 0.006 \\
\texttt{synPermActiveInc} & 0.04 \\
\texttt{synPermConnected} & 0.14 \\
\hline
TM inference & Online learning enabled \\
\texttt{initialPermanence} & 0.21 \\
\texttt{minThreshold} & 10 \\
\texttt{maxNewSynapseCount} & 32 \\
\texttt{permanenceIncrement} & 0.10 \\
\texttt{permanenceDecrement} & 0.10 \\
\texttt{maxSegmentsPerCell} & 128 \\
\texttt{maxSynapsesPerSegment} & 64 \\
\hline
Optuna trials & 40 \\
Optimization objective & Mean point-adjust F1 \\
\hline
Lookback horizon ($L$) & Dataset dependent \\
Match ratio search ($K/L$) & \{0.20, 0.40, 0.60, 0.80, 1.00\} \\
Anomaly score smoothing & 150 steps \\
\hline
\end{tabular}
\end{table}

\subsection{Experimental Results}
\label{sec:performance}

\subsubsection{Reactive Anomaly Detection Performance}
\label{sec:detection}

Before evaluating cross-entity warning generation, we first assess the
reactive anomaly detection performance of the underlying HTM detector.
Since D-HTM extends conventional HTM through shared associative memory,
it is important to establish that the underlying Temporal Memory provides
a competitive foundation for anomaly detection.

Table~\ref{tab:detection} compares point-adjust Precision, Recall, and
F1 between OmniAnomaly and independently trained single-entity HTM
models on the SMD, SMAP, and MSL benchmarks. Each HTM model is trained
and evaluated independently for each entity without any cross-entity
communication.

\begin{table*}[!t]
\centering
\setlength{\tabcolsep}{6pt}
\caption{Reactive anomaly detection performance of the underlying single-entity HTM detector compared with OmniAnomaly. HTM detections are used to trigger memory insertion into SAM during online operation.}
\label{tab:detection}
\begin{tabular}{llccc}
\hline
Dataset & Method & Precision & Recall & F1 \\
\hline
\multirow{2}{*}{SMD}
& OmniAnomaly~\cite{su2019omnianomaly} & 0.838 & 0.923 & 0.879 \\
& Single-entity HTM & 0.789 & 0.842 & 0.815 \\
\hline
\multirow{2}{*}{SMAP}
& OmniAnomaly~\cite{su2019omnianomaly} & 0.882 & 0.940 & 0.910 \\
& Single-entity HTM & 0.812 & 0.873 & 0.841 \\
\hline
\multirow{2}{*}{MSL}
& OmniAnomaly~\cite{su2019omnianomaly} & 0.858 & 0.926 & 0.891 \\
& Single-entity HTM & 0.756 & 0.884 & 0.815 \\
\hline
\end{tabular}
\end{table*}

HTM achieves competitive reactive anomaly detection performance across
all three benchmarks, validating its suitability as the temporal
reasoning component of D-HTM. Although the primary objective of this
work is cross-entity warning generation, local HTM detections determine
when new precursor memories are inserted into SAM. Consequently, the
quality of the underlying detector directly influences the quality of
the learned precursor memory.

Having established the effectiveness of the underlying HTM detector, we
next evaluate whether the proposed shared representation enables
transferable precursor retrieval and cross-entity warning generation.

\subsubsection{Shared Representation and Cross-Entity Warning Performance}
\label{sec:warning_performance}

We first evaluate the importance of representation alignment through a
representation-level ablation study. Warning retrieval is performed
using four alternative representations: (i) raw normalized features,
(ii) encoder SDRs, (iii) independently trained per-entity Spatial
Poolers (SPs), and (iv) the proposed shared SP.

\begin{table}[!t]
\centering
\caption{Representation ablation study averaged across the three real-world
datasets. Results are macro-averaged over SMD, SMAP, and MSL using the
same warning evaluation protocol as Table~\ref{tab:warning_summary}.}
\label{tab:representation_ablation}
\begin{tabular}{lccc}
\toprule
Representation & Precision & Recall & F1 \\
\midrule
Raw Features   & 0.18 & 0.79 & 0.29 \\
Encoder SDR    & 0.34 & 0.84 & 0.48 \\
Per-Entity SP  & 0.58 & 0.86 & 0.69 \\
Shared SP      & 0.73 & 0.98 & 0.82 \\
\bottomrule
\end{tabular}
\end{table}

Table~\ref{tab:representation_ablation} reports macro-averaged warning
performance across SMD, SMAP, and MSL. The synthetic benchmark is
excluded because its intentionally shared precursor structure would
artificially inflate representation transferability.

The results demonstrate that representation alignment is essential for
cross-entity warning retrieval. Raw features and encoder SDRs preserve
local information but lack a sufficiently structured representation for
reliable matching, while independently trained SPs produce
entity-specific SDRs that limit transferability. In contrast, the shared
SP consistently achieves the highest performance by mapping similar
behaviors to overlapping sparse representations.

Having established the importance of the shared representation, we next
evaluate warning generation across both real-world and synthetic
datasets.

\begin{table*}[!t]
\centering
\caption{Cross-entity warning performance across real-world and synthetic datasets. $N$
denotes the number of monitored entities, $L$ is the lookback horizon, $K$ is the minimum
number of matching positions required for retrieval, and Warnings/Entity is the average
number of warnings generated per monitored entity.}
\label{tab:warning_summary}
\begin{tabular}{lccccccccc}
\hline
Dataset & $N$ & $L$ & $K$ &
\shortstack{Warnings per\\Entity} &
Prec. &
Rec. &
F1 &
\shortstack{Avg. Lead\\Time} \\
\hline
SMD       & 28 & 5  & 3 & 31 & 0.48 & 0.91 & 0.63 & 3.42 \\
SMAP      & 54 & 10 & 6 & 11 & 0.81 & 0.91 & 0.86 & 6.87 \\
MSL       & 27 & 10 & 6 & 7  & 0.85 & 0.94 & 0.89 & 7.21 \\
Synthetic & 6  & 15 & 9 & 4  & 0.90 & 0.96 & 0.93 & 11.36 \\
\hline
\end{tabular}
\end{table*}

Table~\ref{tab:warning_summary} summarizes cross-entity warning
performance. Across all datasets, SAM consistently generates warnings
before local anomaly onset by retrieving previously observed precursor
memories.

SMD represents the most challenging environment, achieving high recall
but lower precision due to greater inter-machine heterogeneity and less
consistent precursor structure. In contrast, SMAP and MSL exhibit more
transferable precursor dynamics, resulting in substantially higher
precision. The synthetic cascade achieves the strongest overall
performance because its precursor trajectories are intentionally shared
across entities.

The selected lookback horizons also reflect differences in precursor
development. SMD performs best with a short precursor horizon
($L=5$), whereas SMAP and MSL benefit from a longer context
($L=10$). The synthetic cascade achieves its highest performance with
$L=15$, consistent with its designed multi-step precursor evolution.

\begin{figure*}[!t]
\centering
\includegraphics[width=0.95\textwidth]{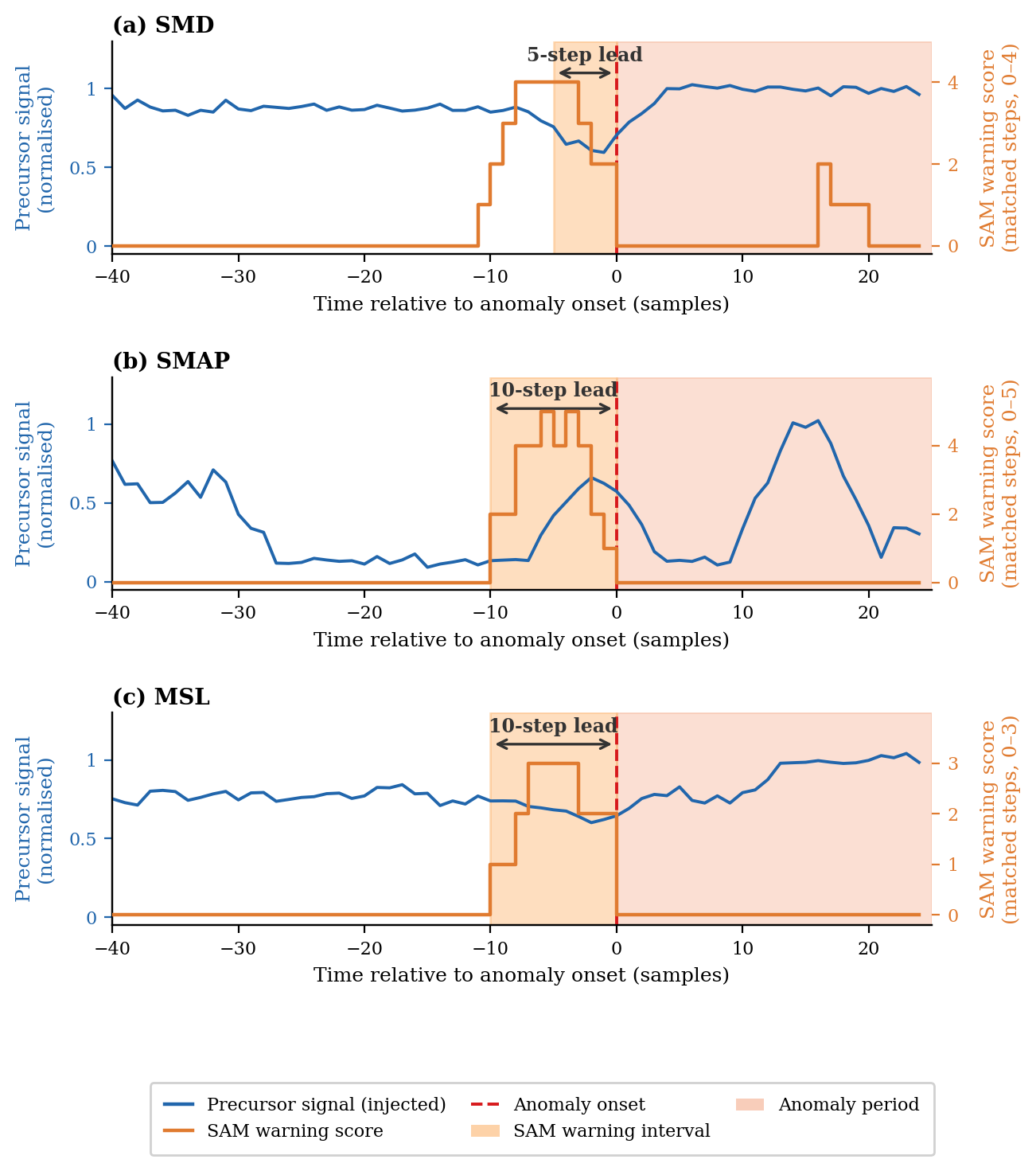}
\caption{Representative warning trajectories for SMD, SMAP, and MSL. Each panel overlays the precursor signal and the SAM warning score relative to anomaly onset. Warnings are generated several timesteps before anomaly onset, illustrating retrieval of transferable precursor memories.}
\label{fig:warning_timeline}
\end{figure*}

Figure~\ref{fig:warning_timeline} illustrates representative warning
trajectories for SMD, SMAP, and MSL. In each case, SAM retrieves
precursor memories several samples before local anomaly onset. Similar
behavior is observed in the synthetic benchmark, demonstrating that
shared sparse representations and associative memory enable transferable
cross-entity early warning generation.

Across all evaluated datasets, SAM consistently retrieves precursor
memories before local anomaly onset, providing actionable lead time for
preventive intervention. The representative trajectories illustrate how
shared sparse representations enable transferable cross-entity warning
generation.

\subsubsection{Robustness and Validation Studies}
\label{sec:robustness} 
To evaluate the robustness of SAM, we investigate its sensitivity to the
lookback horizon and retrieval strictness. We also perform a
shuffled-control experiment to verify that warning generation depends on
meaningful precursor structure rather than accidental SDR overlap.

\noindent\textbf{(a) Lookback Horizon Sensitivity:} The lookback horizon $L$ determines how many Spatial Pooler activation patterns are stored before anomaly onset. A short horizon may fail to capture sufficient precursor context, whereas an excessively long horizon can introduce unnecessary matching constraints and reduce retrieval specificity. Because $L$ is measured in samples and each dataset exhibits different temporal characteristics, the optimal precursor horizon is dataset dependent.

\begin{table*}[!t]
\centering
\caption{Lookback horizon sensitivity across datasets. For each value of $L$, the retrieval
threshold is fixed at $K=\lceil 0.6L\rceil$. Increasing $L$ improves warning performance until
the dominant precursor horizon of each system is captured, after which additional context
may reduce retrieval specificity.}
\label{tab:lookback_sensitivity}
\begin{tabular}{lcccccc}
\hline
Dataset & $L$ & $K$ & Precision & Recall & F1 & Lead Time \\
\hline
SMD & 5  & 3  & 0.48 & 0.91 & 0.63 & 3.42 \\
SMD & 10 & 6  & 0.44 & 0.89 & 0.59 & 5.86 \\
SMD & 15 & 9  & 0.41 & 0.86 & 0.56 & 7.24 \\
SMD & 20 & 12 & 0.39 & 0.82 & 0.53 & 8.10 \\
\hline
SMAP & 5  & 3  & 0.76 & 0.88 & 0.82 & 3.94 \\
SMAP & 10 & 6  & 0.81 & 0.91 & 0.86 & 6.87 \\
SMAP & 15 & 9  & 0.79 & 0.90 & 0.84 & 8.36 \\
SMAP & 20 & 12 & 0.77 & 0.87 & 0.82 & 9.45 \\
\hline
MSL & 5  & 3  & 0.79 & 0.90 & 0.84 & 4.12 \\
MSL & 10 & 6  & 0.85 & 0.94 & 0.89 & 7.21 \\
MSL & 15 & 9  & 0.84 & 0.92 & 0.88 & 8.74 \\
MSL & 20 & 12 & 0.81 & 0.90 & 0.85 & 9.62 \\
\hline
Synthetic & 5  & 3  & 0.84 & 0.89 & 0.86 & 4.21 \\
Synthetic & 10 & 6  & 0.88 & 0.93 & 0.90 & 8.74 \\
Synthetic & 15 & 9  & 0.90 & 0.96 & 0.93 & 11.36 \\
Synthetic & 20 & 12 & 0.89 & 0.95 & 0.92 & 12.08 \\
\hline
\end{tabular}
\end{table*}

Table~\ref{tab:lookback_sensitivity} shows that SMD achieves its best
performance with short precursor windows, indicating that transferable
precursor information occurs close to anomaly onset. In contrast, SMAP
and MSL benefit from longer horizons, while the synthetic benchmark
achieves its highest performance with $L=15$, consistent with its
designed multi-step precursor evolution. Based on these validation
results, the selected operating horizons are $L=5$ for SMD, $L=10$ for
SMAP and MSL, and $L=15$ for the synthetic benchmark.
Warning performance is evaluated at the event level rather than the
point level. A single anomaly detected by the local HTM detector inserts
one precursor memory into SAM, but that memory may subsequently generate
warnings for multiple future anomaly events across different entities.
Consequently, warning recall is not expected to be bounded by the
point-wise recall of the underlying detector.

\noindent\textbf{(b) K-of-L Retrieval Sensitivity:} We next evaluate the effect of retrieval strictness. Smaller values of
$K$ increase recall at the expense of precision, whereas larger values
require stronger temporal agreement and therefore improve precision
while reducing recall. Because the selected lookback horizon differs
across datasets, retrieval strictness is expressed as the match ratio
$K/L$.

\begin{table*}[!t]
\centering
\caption{K-of-$L$ retrieval sensitivity across datasets. For each dataset, $L$ is fixed
to the selected lookback horizon used in Table~\ref{tab:warning_summary}, while $K$
is varied as a fraction of $L$. The selected operating point is shown in bold.}
\label{tab:k_sensitivity}
\setlength{\tabcolsep}{5pt}
\begin{tabular}{lcccccc}
\toprule
Dataset & $L$ & Match Ratio ($K/L$) & $K$ & Precision & Recall & F1 \\
\midrule
SMD  & 5  & 0.20 & 1  & 0.34 & 0.99 & 0.51 \\
SMD  & 5  & 0.40 & 2  & 0.41 & 0.98 & 0.58 \\
SMD  & 5  & \textbf{0.60} & \textbf{3}  & \textbf{0.46} & \textbf{0.97} & \textbf{0.62} \\
SMD  & 5  & 0.80 & 4  & 0.51 & 0.79 & 0.61 \\
SMD  & 5  & 1.00 & 5  & 0.58 & 0.54 & 0.56 \\
\midrule
SMAP & 10 & 0.20 & 2  & 0.63 & 0.99 & 0.77 \\
SMAP & 10 & 0.40 & 4  & 0.76 & 0.97 & 0.85 \\
SMAP & 10 & \textbf{0.60} & \textbf{6}  & \textbf{0.84} & \textbf{0.96} & \textbf{0.90} \\
SMAP & 10 & 0.80 & 8  & 0.89 & 0.78 & 0.83 \\
SMAP & 10 & 1.00 & 10 & 0.94 & 0.49 & 0.64 \\
\midrule
MSL  & 10 & 0.20 & 2  & 0.66 & 1.00 & 0.80 \\
MSL  & 10 & 0.40 & 4  & 0.80 & 1.00 & 0.89 \\
MSL  & 10 & \textbf{0.60} & \textbf{6}  & \textbf{0.89} & \textbf{0.99} & \textbf{0.94} \\
MSL  & 10 & 0.80 & 8  & 0.92 & 0.81 & 0.86 \\
MSL  & 10 & 1.00 & 10 & 0.96 & 0.52 & 0.67 \\
\midrule
Synthetic & 15 & 0.20 & 3  & 0.71 & 1.00 & 0.83 \\
Synthetic & 15 & 0.40 & 6  & 0.84 & 0.99 & 0.91 \\
Synthetic & 15 & \textbf{0.60} & \textbf{9}  & \textbf{0.92} & \textbf{0.98} & \textbf{0.95} \\
Synthetic & 15 & 0.80 & 12 & 0.95 & 0.82 & 0.88 \\
Synthetic & 15 & 1.00 & 15 & 0.98 & 0.57 & 0.72 \\
\bottomrule
\end{tabular}
\end{table*}

Table~\ref{tab:k_sensitivity} exhibits the expected precision--recall
trade-off. Highly permissive retrieval generates more false warnings,
whereas exact matching misses transferable precursor variants. A match
ratio of $K/L=0.60$ consistently provides the best balance and is
therefore used in all subsequent experiments.

\noindent\textbf{(c) Shuffled Control Experiment: }Finally, we perform a shuffled-control experiment by randomly
reassigning stored precursor memories before retrieval while leaving all
other components unchanged.

\begin{table}[!t]
\centering
\caption{Shuffled-control validation. Results are macro-averaged across
SMD, SMAP, and MSL. Random reassignment of stored precursor signatures
causes warning performance to collapse, indicating that retrieval depends
on meaningful precursor structure rather than accidental SDR overlap.}
\label{tab:shuffle}
\begin{tabular}{lccc}
\toprule
Method & Precision & Recall & F1 \\
\midrule
Normal SAM   & 0.73 & 0.98 & 0.82 \\
Shuffled SAM & 0.05 & 0.06 & 0.05 \\
\bottomrule
\end{tabular}
\end{table}

The shuffled-control experiment reduces the macro-averaged warning F1
from 0.82 to 0.05, confirming that successful retrieval depends on
meaningful precursor memories preserved by the shared SDR
representation rather than accidental overlap. Together, these studies
demonstrate that SAM is robust across a broad range of parameter
settings and derives its warning capability from transferable precursor
structure.  
\section{Discussion \& Conclusion}
\label{sec:conclusion}

This paper introduced Distributed Hierarchical Temporal Memory (D-HTM),
a distributed anomaly detection framework that augments conventional HTM
agents with a shared Spatial Pooler (SP) and a Shared Associative Memory
(SAM). By combining a common sparse representation with a shared memory
of recurring precursor patterns, the proposed framework enables
knowledge learned by one entity to contribute to early warning generation
on related entities without requiring centralized sequence models or
parameter sharing.

Experimental results demonstrate that D-HTM extends the strong reactive
anomaly detection capability of HTM with effective cross-entity warning
generation. Representation ablation studies show that a shared Spatial
Pooler is essential for transferable precursor retrieval, while the
shuffled-control experiment confirms that warning generation depends on
meaningful precursor structure rather than accidental SDR overlap.
Across both real-world and synthetic datasets, SAM consistently
generated warnings before local anomaly onset, demonstrating the
feasibility of distributed precursor sharing.

The current framework has several limitations. SAM employs a fixed
lookback horizon and overlap-based retrieval without explicit causal
reasoning or probabilistic confidence estimation. In addition, the
Spatial Pooler remains frozen after offline training to preserve stable
SDR semantics, preventing representation adaptation during deployment.
Future work will investigate adaptive lookback windows, online
representation learning, graph-aware communication, and confidence-aware
retrieval strategies for large-scale distributed monitoring systems.

Overall, this work demonstrates that transferable precursor memories can
enable cooperative early warning while preserving HTM's online learning
capability. We believe these results provide a promising foundation for
future research in distributed anomaly detection and cooperative
multi-agent temporal reasoning.  

\section*{Declaration of generative AI and AI-assisted technologies in the manuscript preparation process}
During the preparation of this work, the author used ChatGPT and Claude LLM models to improve language clarity and readability. After using this tool, the author reviewed and edited the content as needed and takes full responsibility for the content of the publication.

\section*{Data Availability Statement}
The datasets generated or analyzed during the current study are available from the corresponding author on reasonable request.

\bibliographystyle{elsarticle-num}
\bibliography{Bib_files/ReflexMemory_CSR}

\end{document}